\documentclass[11pt]{article}

\usepackage{microtype}
\usepackage{acl2016}
\usepackage{url}

\makeatletter
\newcommand{\@BIBLABEL}{\@emptybiblabel}
\newcommand{\@emptybiblabel}[1]{}
\makeatother
\usepackage[hidelinks]{hyperref}

\usepackage{times}
\usepackage{color}
\usepackage{enumerate}
\usepackage{amsmath}
\usepackage{amssymb}
\usepackage{amsfonts}
\usepackage{multirow}
\usepackage{latexsym}
\usepackage{array}
\usepackage{tikz}
\usepackage{booktabs}

\aclfinalcopy 

\title{A Thorough Examination of the \\ CNN\slash Daily Mail Reading Comprehension Task}

\author{Danqi Chen \and Jason Bolton \and Christopher D. Manning\\
            Computer Science
	    Stanford University\\
	    Stanford, CA 94305-9020, USA\\
	    {\tt \{danqi,jebolton,manning\}@cs.stanford.edu}}

\DeclareMathOperator*{\argmax}{arg\,max}

\DeclareMathOperator*{\softmax}{softmax}

\newcommand{\mf}[1]{\mathbf{#1}}
\newcommand{\tf}[1]{\textbf{#1}}
\newcommand{\ti}[1]{\textit{#1}}

\newcommand\R{\mathbb{R}}

\newcommand{\finalcnn}{73.6\%}
\newcommand{\finaldm}{76.6\%}

\newcolumntype{C}[1]{>{\centering\let\newline\\\arraybackslash\hspace{0pt}}m{#1}}
\newcommand{\figref}[1]{Figure~\ref{fig:#1}}
\date{}

\begin{document}
\def\tinyweeny{\fontsize{6pt}{8pt}\selectfont}

\maketitle

\begin{abstract}

Enabling a computer to understand a document so that it can answer comprehension questions is a central, yet unsolved goal of NLP\@. A key factor impeding its solution by machine learned systems is the limited availability of human-annotated data. \newcite{hermann2015teaching} seek to solve this problem by creating over a million training examples by pairing \ti{CNN} and \ti{Daily Mail} news articles with their summarized bullet points, and show that a neural network can then be trained to give good performance on this task. In this paper, we conduct a thorough examination of this new reading comprehension task. Our primary aim is to understand what depth of language understanding is required to do well on this task. We approach this from one side by doing a careful hand-analysis of a small subset of the problems and from the other by showing that simple, carefully designed systems can obtain accuracies of {\finalcnn} and {\finaldm} on these two datasets, exceeding current state-of-the-art results by {7}--{10}\%  and approaching what we believe is the ceiling for performance on this task.\footnote{Our code is available at \url{https://github.com/danqi/rc-cnn-dailymail}.}
\end{abstract}

\section{Introduction}

Reading comprehension (RC) is the ability to read text, process it, and understand its meaning.\footnote{\url{https://en.wikipedia.org/wiki/Reading_comprehension}} How to endow computers with this capacity has been an elusive challenge and a long-standing goal of Artificial Intelligence (e.g., \cite{norvig87phd}). Genuine reading comprehension involves interpretation of the text and making complex inferences. Human reading comprehension is often tested by asking questions that require interpretive understanding of a passage, and the same approach has been suggested for testing computers \cite{burges2013towards}.

In recent years, there have been several strands of work which attempt to collect human-labeled data for this task -- in the form of document, question and answer triples --  and to learn machine learning models directly from it \cite{richardson2013mctest,berant2014modeling,wang2015machine}. However, these datasets consist of only hundreds of documents, as the labeled examples usually require considerable expertise and neat design, making the annotation process quite expensive.  The subsequent scarcity of labeled examples prevents us from training powerful statistical models, such as deep learning models, and would seem to prevent a system from learning complex textual reasoning capacities.

\begin{figure}
\centering
\includegraphics[scale=0.38]{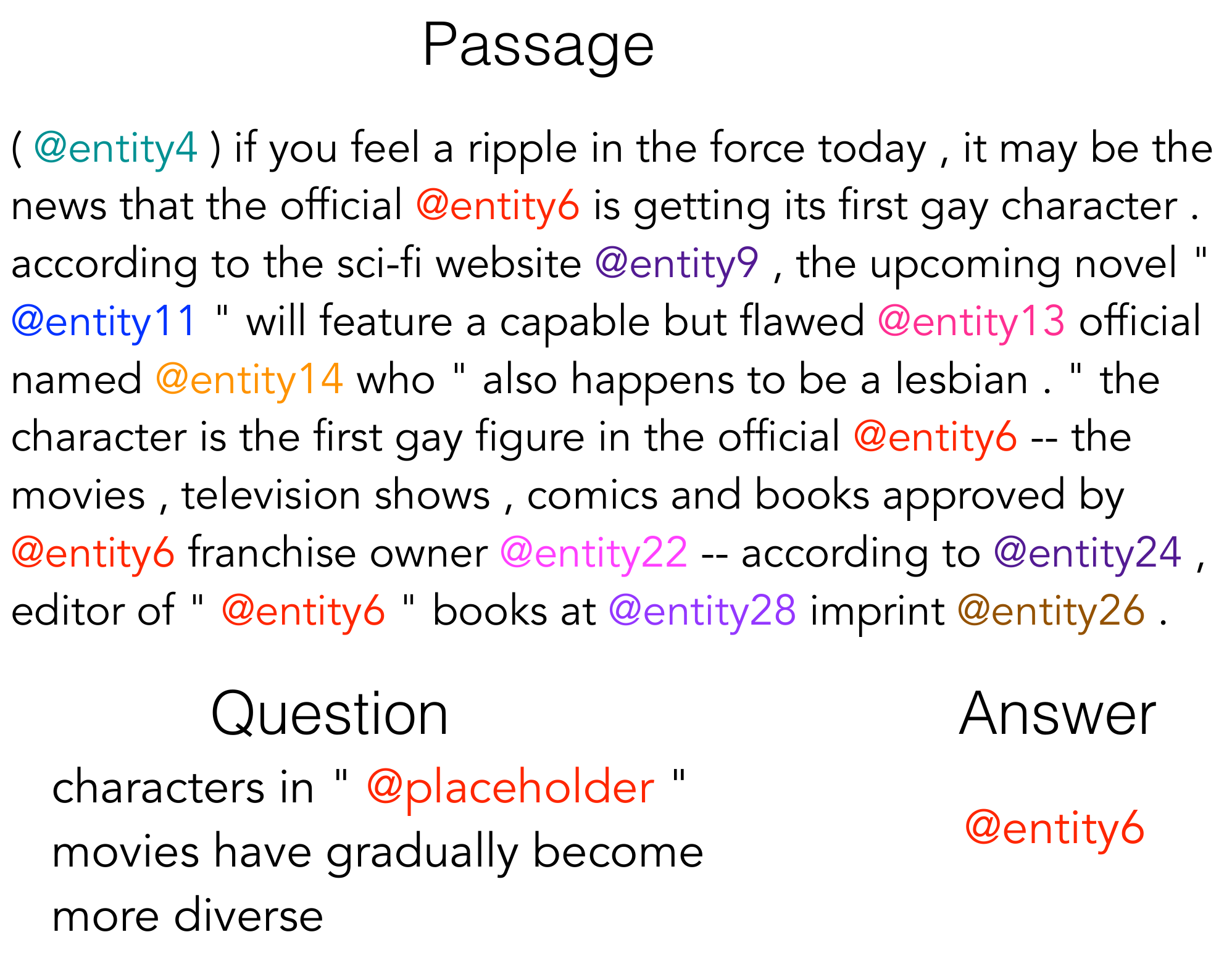}
\vspace{-2em}
\caption{An example item from dataset \ti{CNN}.}
\label{fig:example}
\end{figure}

Recently, researchers at \ti{DeepMind} \cite{hermann2015teaching} had the appealing, original idea of exploiting the fact that the abundant news articles of \ti{CNN} and \ti{Daily Mail} are accompanied by bullet point summaries in order to heuristically create large-scale supervised training data for the reading comprehension task. Figure~\ref{fig:example} gives an example. Their idea is that a bullet point usually summarizes one or several aspects of the article. If the computer understands the content of the article, it should be able to infer the missing entity in the bullet point.

This is a clever way of creating supervised data cheaply and holds promise for making progress on training RC models; however, it is unclear what level of reading comprehension is actually needed to solve this somewhat artificial task and, indeed, what statistical models that do reasonably well on this task have actually learned.

In this paper, our aim is to provide an in-depth and thoughtful analysis of this dataset and what level of natural language understanding is needed to do well on it. We demonstrate that simple, carefully designed systems can obtain high, state-of-the-art accuracies of  {\finalcnn} and {\finaldm} on \ti{CNN} and \ti{Daily Mail} respectively. We do a careful hand-analysis of a small subset of the problems to provide data on their difficulty and what kinds of language understanding are needed to be successful and we try to diagnose what is learned by the systems that we have built.  We conclude that: (i)~this dataset is easier than previously realized, (ii)~straightforward, conventional NLP systems can do much better on it than previously suggested, (iii)~the distributed representations of deep learning systems are very effective at recognizing paraphrases, (iv)~partly because of the nature of the questions, current systems much more have the nature of single-sentence relation extraction systems than larger-discourse-context text understanding systems, (v) the systems that we present here are close to the ceiling of performance for single-sentence and unambiguous cases of this dataset, and (vi)~the prospects for getting the final 20\% of questions correct appear poor, since most of them involve issues in the data preparation which undermine the chances of answering the question (coreference errors or anonymization of entities making understanding too difficult).

\section{The Reading Comprehension Task}


The RC datasets introduced in \cite{hermann2015teaching} are made from articles on the news websites \ti{CNN} and \ti{Daily Mail}, utilizing articles and their bullet point summaries.\footnote{The datasets are available at \url{https://github.com/deepmind/rc-data}.} Figure~\ref{fig:example} demonstrates an example\footnote{The original article can be found at \url{http://www.cnn.com/2015/03/10/entertainment/feat-star-wars-gay-character/}.}: it consists of a passage $p$, a question $q$ and an answer $a$, where the passage is a news article, the question is a cloze-style task, in which one of the article's bullet points has had one entity replaced by a placeholder, and the answer is this questioned entity. The goal is to infer the missing entity (answer $a$) from all the possible entities which appear in the passage. A news article is usually associated with a few (e.g., 3--5) bullet points and each of them highlights one aspect of its content.

The text has been run through a Google NLP pipeline. It it tokenized, lowercased, and named entity recognition and coreference resolution have been run. For each coreference chain containing at least one named entity, all items in the chain are replaced by an @entity$n$ marker, for a distinct index $n$. \newcite{hermann2015teaching} argue convincingly that such a strategy is necessary to ensure that systems approach this task by understanding the passage in front of them, rather than by using world knowledge or a language model to answer questions without needing to understand the passage. However, this also gives the task a somewhat artificial character. On the one hand, systems are greatly helped by entity recognition and coreference having already been performed; on the other, they suffer when either of these modules fail, as they do (in \figref{example}, ``the character'' should probably be coreferent with @entity14; clearer examples of failure appear later on in our data analysis). Moreover, this inability to use world knowledge also makes it much more difficult for a human to do this task -- occasionally it is very difficult or impossible for a human to determine the correct answer when presented with an item anonymized in this way.

\begin{table}
\centering
\begin{tabular}{@{} l r  r @{}}
\toprule
& \tf{CNN} & \tf{Daily Mail} \\
\hline
\# Train & 380,298 & 879,450 \\
\# Dev & 3,924 & 64,835 \\
\# Test & 3,198 & 53,182 \\
\midrule
Passage: avg.\ tokens & 761.8 & 813.1 \\
Passage: avg.\ sentences & 32.3 & 28.9 \\
Question: avg.\ tokens & 12.5 & 14.3 \\
\hline
Avg. \# entities & 26.2 & 26.2 \\
\bottomrule
\end{tabular}
\caption{Data statistics of the \ti{CNN} and \ti{Daily Mail} datasets. The avg.\ tokens and sentences in the passage, the avg.\ tokens in the query, and the number of entities are based on statistics from the training set, but they are similar on the development and test sets.}
\label{table:data_stat}
\end{table}

The creation of the datasets benefits from the sheer volume of news articles available online, so they offer a large and realistic testing ground for statistical models. Table~\ref{table:data_stat} provides some statistics on the two datasets: there are 380k and 879k training examples for \ti{CNN} and \ti{Daily Mail} respectively. The passages are around 30 sentences and 800 tokens on average, while each question contains around 12--14 tokens.

In the following sections, we seek to more deeply understand the nature of this dataset. We first build some straightforward systems in order to get a better idea of a lower-bound for the performance of current NLP systems. Then we turn to data analysis of a sample of the items to examine their nature and an upper bound on performance.

\section{Our Systems}

In this section, we describe two systems we implemented -- a conventional entity-centric classifier and an end-to-end neural network. While \newcite{hermann2015teaching} do provide several baselines for performance on the RC task, we suspect that their baselines are not that strong. They attempt to use a frame-semantic parser, and we feel that the poor coverage of that parser undermines the results, and is not representative of what a straightforward NLP system -- based on standard approaches to factoid question answering and relation extraction developed over the last 15 years -- can achieve. Indeed, their frame-semantic model is markedly inferior to another baseline they provide, a heuristic word distance model. At present just two papers are available presenting results on this RC task, both presenting neural network approaches: \cite{hermann2015teaching} and \cite{hill2016goldilocks}. While the latter is wrapped in the language of end-to-end memory networks, it actually presents a fairly simple window-based neural network classifier running on the CNN data. Its success again raises questions about the true nature and complexity of the RC task provided by this dataset, which we seek to clarify by building a simple attention-based neural net classifier.

Given the (passage, question, answer) triple $(p, q, a)$, $p = \{p_1, \ldots, p_{m}\}$ and $q = \{q_1, \ldots, q_{l}\}$ are sequences of tokens for the passage and question sentence, with $q$ containing exactly one ``@placeholder'' token. The goal is to infer the correct entity $a \in p \cap E$ that the placeholder corresponds to, where $E$ is the set of all abstract entity markers. Note that the correct answer entity must appear in the passage $p$.

\subsection{Entity-Centric Classifier}

We first build a conventional feature-based classifier, aiming to explore what features are effective for this task. This is similar in spirit to \cite{wang2015machine}, which at present has very competitive performance on the MCTest RC dataset \cite{richardson2013mctest}. The setup of this system is to design a feature vector $f_{p, q}(e)$ for each candidate entity $e$, and to learn a weight vector $\theta$ such that the correct answer $a$ is expected to rank higher than all other candidate entities:
\begin{equation}
\theta^{\intercal}f_{p, q}(a) > \theta^{\intercal}f_{p, q}(e), \forall e \in E \cap p \setminus \{a\}
\end{equation}

We employ the following feature templates:
\begin{enumerate}[1.]
    \setlength\itemsep{-0.1em}
    \item
        Whether entity $e$ occurs in the passage.
    \item
        Whether entity $e$ occurs in the question.
    \item
        The frequency of entity $e$ in the passage.
    \item
        The first position of occurence of entity $e$ in the passage.
    \item
        $n$-gram exact match: whether there is an exact match between the text surrounding the placeholder and the text surrounding entity $e$. We have features for all combinations of matching left and/or right one or two words.
    \item
        Word distance: we align the placeholder with each occurrence of entity $e$, and compute the average minimum distance of each non-stop question word from the entity in the passage.
    \item
        Sentence co-occurrence: whether entity $e$ co-occurs with another entity or verb that appears in the question, in some sentence of the passage.
    \item
        Dependency parse match: we dependency parse both the question and all the sentences in the passage, and extract an indicator feature of whether $w \xrightarrow{r} \text{@placeholder}$ and $w \xrightarrow{r} e$ are both found; similar features are constructed for $\text{@placeholder} \xrightarrow{r} w$ and $e \xrightarrow{r} w$.
\end{enumerate}

\subsection{End-to-end Neural Network}

Our neural network system is based on the \ti{AttentiveReader} model proposed by \cite{hermann2015teaching}. The framework can be described in the following three steps (see Figure \ref{fig:framework}):

\begin{figure*}[!ht]
\centering
    \includegraphics[scale=0.37]{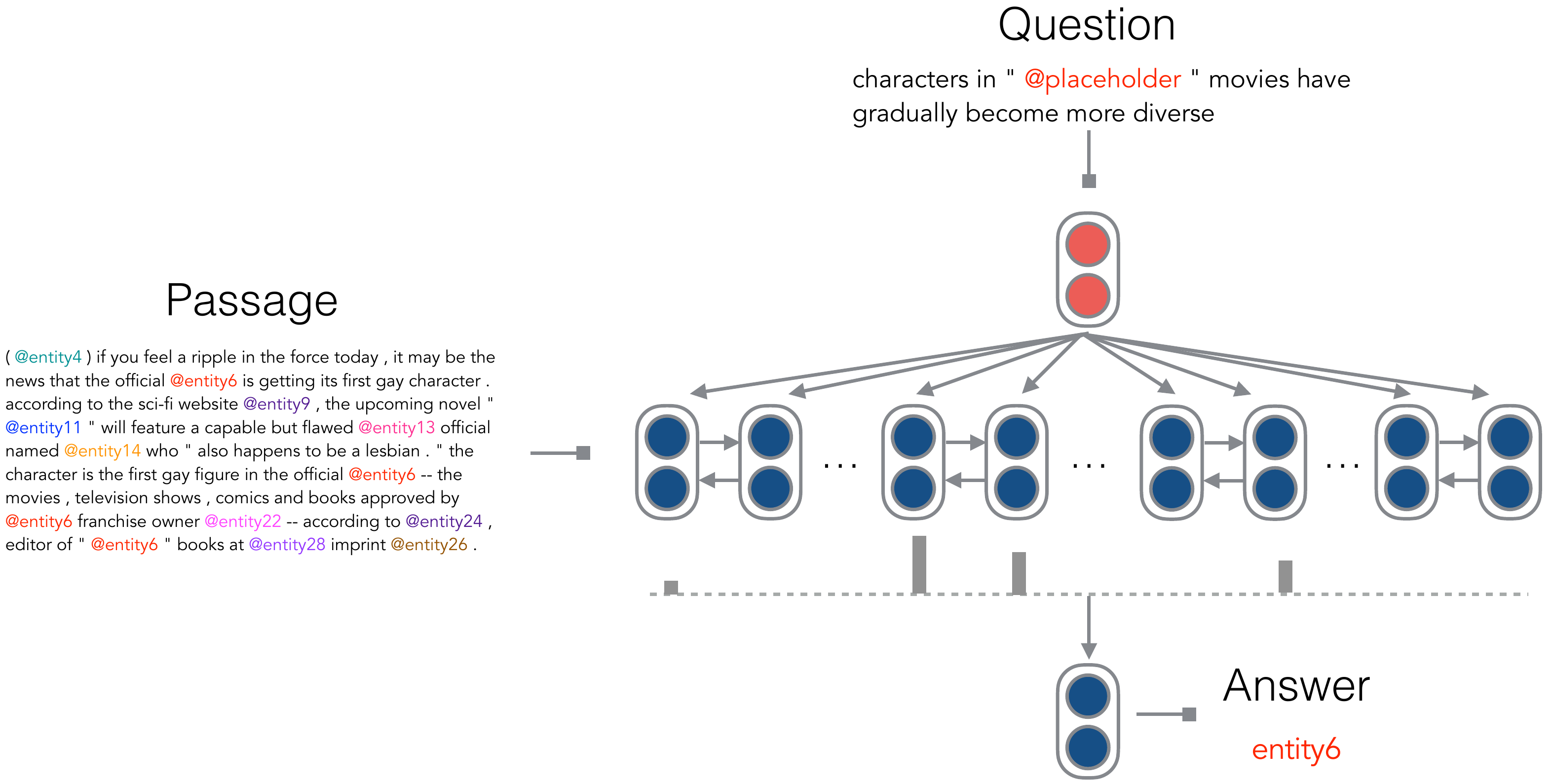}
\caption{Our neural network architecture for the reading comprehension task.}
\label{fig:framework}
\end{figure*}

\begin{description}
    \item[\tf{Encoding:}] First, all the words are mapped to $d$-dimensional vectors via an embedding matrix $E \in \R^{d \times |\mathcal{V}|}$; therefore we have $p$: $\mf{p}_1, \ldots, \mf{p}_m \in R^d$ and $q: \mf{q}_1, \ldots, \mf{q}_{l} \in R^d$.

        Next we use a shallow bi-directional recurrent neural network (RNN) with hidden size $\tilde{h}$ to encode contextual embeddings $\tilde{\mf{p}}_i$ of each word in the passage,
        \begin{eqnarray*}
            \overrightarrow{\mf{h}}_i & = & \text{RNN}(\overrightarrow{\mf{h}}_{i-1}, \mf{p}_i), i = 1, \ldots, m\\
            \overleftarrow{\mf{h}}_i & = & \text{RNN}(\overleftarrow{\mf{h}}_{i+1}, \mf{p}_i), i = m, \ldots, 1
        \end{eqnarray*}
        and $\tilde{\mf{p}}_i = \text{concat}(\overrightarrow{\mf{h}}_i, \overleftarrow{\mf{h}}_i) \in \R^{h}$, where $h = 2 \tilde{h}$.
        Meanwhile, we use another bi-directional RNN to map the question $\mf{q}_1, \ldots, \mf{q}_l$ to an embedding $\mf{q} \in \R^h$. We choose to use Gated Recurrent Unit (GRU) \cite{cho2014learning} in our experiments because it performs similarly but is computationally cheaper than LSTM.

    \item[\tf{Attention:}] In this step, the goal is to compare the question embedding and all the contextual embeddings, and \ti{select} the pieces of information that are relevant to the question. We compute a probability distribution $\alpha$ depending on the degree of relevance between word $p_i$ (in its context) and the question $q$ and then produce an output vector $\mf{o}$ which is a weighted combination of all contextual embeddings $\{\tilde{\mf{p}}_i\}$:
        \begin{eqnarray}
            \alpha_i & = & \softmax\nolimits_i \mf{q} ^{\intercal} \mf{W}_{s} \tilde{\mf{p}}_i  \\
            \mf{o} & = & \sum\nolimits_{i}{\alpha_i \tilde{\mf{p}}_i}
        \end{eqnarray}
        $\mf{W_s} \in \R^{h \times h}$ is used in a bilinear term, which allows us to compute a similarity between $\mf{q}$ and $\tilde{\mf{p}}_i$ more flexibly than with just a dot product.
    \item[\tf{Prediction:}] Using the \ti{output} vector $\mf{o}$, the system outputs the most likely answer using:
        \begin{equation}
            a = \argmax\nolimits_{a \in p \cap E}{W_a ^{\intercal} \mf{o}}
        \end{equation}
        Finally, the system adds a softmax function on top of $W_a ^{\intercal} \mf{o}$ and adopts a negative log-likelihood objective for training.
\end{description}

\paragraph*{Differences from \cite{hermann2015teaching}.}
Our model basically follows the \ti{AttentiveReader}. However, to our surprise, our experiments observed nearly \tf{7} --\tf{10}\% improvement over the original \ti{AttentiveReader} results on \ti{CNN} and \ti{Daily Mail} datasets (discussed in Sec.~\ref{sec:experiments}). Concretely, our model has the following differences:
\begin{itemize}
\item
We use a bilinear term, instead of a $\tanh$ layer to compute the relevance (attention) between question and contextual embeddings. The effectiveness of the simple bilinear attention function has been shown previously for neural machine translation by \cite{luong2015effective}.
\item
After obtaining the weighted contextual embeddings $\mf{o}$, we use $\mf{o}$ for direct prediction. In contrast, the original model in \cite{hermann2015teaching} combined $\mf{o}$ and the question embedding $\mf{q}$ via another non-linear layer before making final predictions. We found that we could remove this layer without harming performance. We believe it is sufficient for the model to learn to return the entity to which it maximally gives attention.
\item
The original model considers all the words from the vocabulary $\mathcal{V}$ in making predictions. We think this is unnecessary, and only predict among entities which appear in the passage.
\end{itemize}
Of these changes, only the first seems important; the other two just aim at keeping the model simple.

\paragraph*{Window-based MemN2Ns \cite{hill2016goldilocks}.}
Another recent neural network approach proposed by \cite{hill2016goldilocks} is based on a memory network architecture \cite{weston2015memory}. We think it is highly similar in spirit. The biggest difference is their way of encoding passages: they demonstrate that it is most effective to only use a 5-word context window when evaluating a candidate entity and they use a positional unigram approach to encode the contextual embeddings: if a window consists of 5 words $x_1, \ldots, x_5$, then it is encoded as $\sum_{i=1}^{5}{E_i(x_i)}$, resulting in $5$ separate embedding matrices to learn. They encode the 5-word window surrounding the placeholder in a similar way and all other words in the question text are ignored. In addition, they simply use a dot product to compute the ``relevance'' between the question and a contextual embedding. This simple model nevertheless works well, showing the extent to which this RC task can be done by very local context matching.

\section{Experiments}
\label{sec:experiments}
\subsection{Training Details}
For training our conventional classifier, we use the implementation of \ti{LambdaMART} \cite{wu2010adapting} in the RankLib package.\footnote{\url{https://sourceforge.net/p/lemur/wiki/RankLib/}.} We use this ranking algorithm since our problem is naturally a ranking problem and forests of boosted decision trees have been very successful lately (as seen, e.g., in many recent Kaggle competitions). We do not use all the features of \ti{LambdaMART} since we are only scoring 1/0 loss on the first ranked proposal, rather than using an IR-style metric to score ranked results. We use Stanford's neural network dependency parser \cite{chen2014fast} to parse all our document and question text, and all other features can be extracted without additional tools.

\looseness=-1
For training our neural networks, we only keep the most frequent $|\mathcal{V}| = 50\text{k}$ words (including entity and placeholder markers), and map all other words to an \ti{<unk>} token. We choose word embedding size $d = 100$, and use the $100$-dimensional pre-trained \ti{GloVe} word embeddings \cite{pennington2014glove} for initialization. The attention and output parameters are initialized from a uniform distribution between $(-0.01, 0.01)$, and the GRU weights are initialized from a Gaussian distribution $\mathcal{N}(0, 0.1)$.

We use hidden size $h = 128$ for \ti{CNN} and 256 for \ti{Daily Mail}. Optimization is carried out using vanilla stochastic gradient descent (SGD), with a fixed learning rate of $0.1$. We sort all the examples by the length of its passage, and randomly sample a mini-batch of size 32 for each update. We also apply dropout with probability $0.2$ to the embedding layer and gradient clipping when the norm of gradients exceeds $10$.

Additionally, we think the original indices of entity markers are generated arbitrarily. We attempt to relabel the entity markers based on their first occurrence in the passage and question \footnote{The first occurring entity is relabeled as @entity1, and the second one is relabeled as @entity2, and so on.} and find that this step can make training converge faster as well bring slight gains. We report both results (with and without relabeling) for future reference.

All of our models are run on a single GPU (GeForce GTX TITAN X), with roughly a runtime of 3 hours per epoch for \ti{CNN}, and 12 hours per epoch for \ti{Daily Mail}. We run all the models up to $30$ epochs and select the model that achieves the best accuracy on the development set.

We run our models 5 times independently with different random seeds and report average performance across the runs. We also report ensemble results which average the prediction probabilities of the 5 models.



\subsection{Main Results}

\begin{table*}
\centering
\begin{tabular}{l C{1.5cm} C{1.5cm} C{1.5cm} C{1.5cm}}
\toprule
\multirow{2}{*}{Model} & \multicolumn{2}{c}{CNN} &  \multicolumn{2}{c}{Daily Mail} \\
& Dev & Test & Dev & Test \\
\midrule
 Frame-semantic model $^\dagger$ &36.3  & 40.2 & 35.5 & 35.5 \\
 Word distance model $^\dagger$ & 50.5 & 50.9 & 56.4 & 55.5 \\
 Deep LSTM Reader $^\dagger$ & 55.0 & 57.0 & 63.3 & 62.2 \\
Attentive Reader $^\dagger$ & 61.6 & 63.0 & 70.5 & 69.0 \\
 Impatient Reader $^\dagger$ & 61.8 & 63.8 & 69.0 & 68.0 \\
\midrule
MemNNs (window memory) $^\ddagger$ & 58.0 & 60.6 & N/A & N/A \\
MemNNs (window memory + self-sup.) $^\ddagger$ & 63.4 & 66.8 & N/A & N/A\\
MemNNs (ensemble) $^\ddagger$ & 66.2\rlap{$^*$} & 69.4\rlap{$^*$} & N/A & N/A \\
\midrule
Ours: Classifier & 67.1 & 67.9 & 69.1 & 68.3 \\
\midrule
Ours: Neural net & 72.5 & 72.7 & 76.9 & 76.0 \\
Ours: Neural net (ensemble) &  76.2\rlap{$^*$} & 76.5\rlap{$^*$} & 79.5\rlap{$^*$} & 78.7\rlap{$^*$} \\
\midrule
Ours: Neural net (relabeling) &  \tf{73.8} & \tf{73.6} & \tf{77.6} & \tf{76.6} \\
Ours: Neural net (relabeling, ensemble) & \tf{77.2}\rlap{$^*$} & \tf{77.6}\rlap{$^*$} & \tf{80.2}\rlap{$^*$} & \tf{79.2}\rlap{$^*$}\\
\bottomrule
\end{tabular}
\caption{Accuracy of all models on the \ti{CNN} and \ti{Daily Mail} datasets. Results marked $^\dagger$ are from \protect\cite{hermann2015teaching} and results marked $^\ddagger$ are from \protect\cite{hill2016goldilocks}. \ti{Classifier} and \ti{Neural net} denote our entity-centric classifier and neural network systems respectively. The numbers marked with $^*$ indicate that the results are from ensemble models.}
\label{table:main-results}
\end{table*}

Table~\ref{table:main-results} presents our main results. The conventional feature-based classifier obtains $67.9\%$ accuracy on the \ti{CNN} test set. Not only does this significantly outperform any of the symbolic approaches reported in \cite{hermann2015teaching}, it also outperforms all the neural network systems from their paper and the best single-system result reported so far from \cite{hill2016goldilocks}. This suggests that the task might not be as difficult as suggested, and a simple feature set can cover many of the cases. Table \ref{table:feature-ablation} presents a feature ablation analysis of our entity-centric classifier on the development portion of the \ti{CNN} dataset. It shows that $n$-gram match and frequency of entities
are the two most important classes of features.

\begin{table}
\begin{center}
\begin{tabular}{l  c }
\toprule
Features & Accuracy \\
\midrule
Full model &  67.1 \\
$-$ whether $e$ is in the passage & 67.1 \\
$-$ whether $e$ is in the question & 67.0 \\
$-$ frequency of $e$ & \tf{63.7} \\
$-$ position of $e$ & 65.9 \\
$-$ $n$-gram match & \tf{60.5} \\
$-$ word distance & 65.4 \\
$-$ sentence co-occurrence & 66.0 \\
$-$ dependency parse match & 65.6 \\
\bottomrule
\end{tabular}
\end{center}
\caption{Feature ablation analysis of our entity-centric classifier on
  the development portion of the \ti{CNN} dataset. The numbers denote
  the accuracy after we exclude each feature from the full system, so
  a low number indicates an important feature.}
\label{table:feature-ablation}
\end{table}

More dramatically, our single-model neural network surpasses the previous results by a large margin (over 5\%). The relabeling process further improves the results by $0.6\%$ and $0.9\%$, pushing up the state-of-the-art accuracies to {\finalcnn} and {\finaldm} on the two datasets respectively. The ensembles of $5$ models consistently bring further $2 - 4\%$ gains.




Concurrently with our paper, \newcite{kadlec2016text} and \newcite{kobayashi2016dynamic} also experiment on these two datasets and report competitive results. However, our model not only still outperforms theirs, but also appears to be structurally simpler. All these recent efforts converge to similar numbers, and we believe that they are approaching the ceiling performance of this task, as we will indicate in the next section.


\section{Data Analysis}
\label{sec:data-analysis}
So far, we have good results via either of our systems. In this section, we aim to conduct an in-depth analysis and answer the following questions: (i)~Since the dataset was created in an automatic and heuristic way, how many of the questions are trivial to answer, and how many are noisy and not answerable? (ii)~What have these models learned? What are the prospects for further improving them?
To study this, we randomly sampled 100 examples from the dev portion of the \ti{CNN} dataset for analysis (see more details in Appendix \ref{appendix:samples}).


\subsection{Breakdown of the Examples}

\begin{table*}
\centering
\begin{tabular}{@{} p{1.45cm}  p{5.2cm} p{8.5cm} @{}}
\toprule
Category & Question & Passage \\
\midrule
Exact Match & \ti{it 's clear @entity0 is leaning toward} {\tf{@placeholder}} ,  says an expert who monitors @entity0 & \ldots @entity116 , who follows @entity0 's operations and propaganda closely , recently told @entity3 , \ti{it 's clear @entity0 is leaning toward} \tf{@entity60}  in terms of doctrine , ideology and an emphasis on holding territory after operations . \ldots  \\
\midrule
Para-phrase & {\tf{@placeholder} says he understands why @entity0 wo n't play at his tournament} &  \ldots @entity0 called me personally to let me know that he would n't be playing here at @entity23 , " \tf{@entity3} said on his @entity21 event 's website . \ldots \\
\midrule
Partial clue & a tv movie based on @entity2 's book \tf{@placeholder} casts a @entity76 actor as @entity5 & \ldots  to @entity12  @entity2 professed that his \tf{@entity11} is not a religious book . \ldots \\
\midrule
Multiple sent. &  he 's doing a his - and - her duet all by himself ,  @entity6 said of \tf{@placeholder} &  \ldots we got some groundbreaking performances , here too , tonight ,  @entity6 said . we got \tf{@entity17} , who will be doing some musical performances . he 's doing a his - and - her duet all by himself .  \ldots \\
\midrule
Coref. Error & rapper \tf{@placeholder} " disgusted , " cancels upcoming show for @entity280 & \ldots with hip - hop star \tf{@entity246} saying on @entity247 that he was canceling an upcoming show for the @entity249 . \ldots  (but @entity249 = @entity280 = SAEs)\\
\midrule
Hard & pilot error and snow were reasons stated for \tf{@placeholder} plane crash  & \ldots a small aircraft carrying \tf{@entity5} , @entity6 and @entity7 the @entity12  @entity3 crashed a few miles from @entity9 , near @entity10 , @entity11 . \ldots \\
\bottomrule
\end{tabular}
\caption{Some representative examples from each category. }
\label{table:category-examples}
\end{table*}

 After carefully analyzing these 100 examples,  we roughly classify them into the following categories (if an example satisfies more than one category, we classify it into the earliest one):
\begin{description}
    \item[\tf{Exact match}] The nearest words around the placeholder are also found in the passage surrounding an entity marker; the answer is self-evident.
    \item[\tf{Sentence-level paraphrasing}] The question text is entailed\slash rephrased by \ti{exactly} one sentence in the passage, so the answer can definitely be identified from that sentence.
    \item[\tf{Partial clue}] In many cases, even though we cannot find a complete semantic match between the question text and some sentence, we are still able to infer the answer through partial clues, such as some word/concept overlap.
    \item[\tf{Multiple sentences}] It requires processing multiple sentences to infer the correct answer.
    \item[\tf{Coreference errors}] It is unavoidable that there are many coreference errors in the dataset. This category includes those examples with critical coreference errors for the answer entity or key entities appearing in the question. Basically we treat this category as ``not answerable''.
    \item[\tf{Ambiguous or very hard}] This category includes examples for which we think humans are not able to obtain the correct answer (confidently).
\end{description}

\begin{table}
  \centering
    \begin{tabular}{l  l  r}
      \toprule
    No. & Category &  (\%)  \\
    \midrule
    1 & Exact match & 13   \\
    2 & Paraphrasing & 41 \\
    3 & Partial clue & 19  \\
    4 & Multiple sentences & 2  \\
    \midrule
    5 & Coreference errors & 8 \\
    6 & Ambiguous / hard &  17 \\
    \bottomrule
    \end{tabular}
    \caption{An estimate of the breakdown of the dataset into classes,
      based on the analysis of our sampled 100 examples from the \ti{CNN} dataset.}
    \label{table:example-breakdown}
\end{table}

Table~\ref{table:example-breakdown} provides our estimate of the percentage for each category, and Table \ref{table:category-examples} presents one representative example from each category.
To our surprise, ``coreference errors'' and ``ambiguous\slash hard'' cases account for $25\%$ of this sample set, based on our manual analysis, and this certainly will be a barrier  for training models with an accuracy much above 75\% (although, of course, a model can sometimes make a lucky guess). Additionally, only 2 examples require multiple sentences for inference -- this is a lower rate than  we expected and \newcite{hermann2015teaching} suggest. Therefore, we hypothesize that in most of the ``answerable'' cases, the goal is to identify the most relevant (single) sentence, and then to infer the answer based upon it.

\subsection{Per-category Performance}

Now,
we further analyze the predictions of our two systems, based on the above categorization.

\begin{table}
   \centering
    \begin{tabular}{@{} l  r @{\hspace*{0.25em}} r r @{\hspace*{0.25em}} r @{}}
\toprule
     {Category} &  \multicolumn{2}{c}{{Classifier}} & \multicolumn{2}{c}{{Neural net}} \\
    \midrule
     Exact match & 13 & (100.0\%) & 13 & (100.0\%) \\
     Paraphrasing &  32 & (78.1\%) & 39 & (95.1\%) \\
     Partial clue & 14 & (73.7\%) &  17 & (89.5\%) \\
     Multiple sentences &  1 & (50.0\%) & 1 & (50.0\%) \\
    \midrule
     Coreference errors &  4 & (50.0\%) & 3 & (37.5\%)\\
     Ambiguous / hard &  2 & (11.8\%) & 1 & (5.9\%)  \\
     \midrule
     All & 66 & (66.0\%) & 74 & (74.0\%) \\
    \bottomrule
    \end{tabular}
    \caption{The per-category performance of our two systems.}
    \label{table:category-performance}
\end{table}

As seen in Table \ref{table:category-performance}, we have the following observations: (i)~The exact-match cases are quite simple and both systems get 100\% correct. (ii)~For the ambiguous\slash hard and entity-linking-error cases, meeting our expectations, both of the systems perform poorly. (iii)~The two systems mainly differ in paraphrasing cases, and some of the ``partial clue'' cases. This clearly shows how neural networks are better capable of learning semantic matches involving paraphrasing or lexical variation between the two sentences. (iv)~We believe that the neural-net system already achieves near-optimal performance on all the single-sentence and unambiguous cases. There does not seem to be much useful headroom for exploring more sophisticated natural language understanding approaches on this dataset.


\section{Related Tasks}

We briefly survey other tasks related to reading comprehension.

\tf{MCTest} \cite{richardson2013mctest} is an open-domain reading comprehension task, in the form of fictional short stories, accompanied by multiple-choice questions. It was carefully created using crowd sourcing, and aims at a 7-year-old reading comprehension level.

On the one hand, this dataset has a high demand on various reasoning capacities: over $50\%$ of the questions require multiple sentences to answer and also the questions come in assorted categories (\ti{what}, \ti{why}, \ti{how}, \ti{whose}, \ti{which}, etc). On the other hand, the full dataset has only 660 paragraphs in total (each paragraph is associated with 4 questions), which renders training statistical models (especially complex ones) very difficult.

Up to now, the best solutions \cite{sachan2015learning,wang2015machine} are still heavily relying on manually curated syntactic\slash semantic features, with the aid of additional knowledge (e.g., word embeddings, lexical\slash paragraph databases).

\tf{Children Book Test} \cite{hill2016goldilocks} was developed in a similar spirit to the \ti{CNN}\slash\ti{Daily Mail} datasets. It takes any consecutive 21 sentences from a children's book -- the first 20 sentences are used as the passage, and the goal is to infer a missing word in the 21st sentence (question and answer). The questions are also categorized by the type of the missing word: named entity, common noun, preposition or verb. According to the first study on this dataset \cite{hill2016goldilocks}, a language model (an $n$-gram model or a recurrent neural network) with local context is sufficient for predicting verbs or prepositions; however, for named entities or common nouns, it improves performance to scan through the whole paragraph to make predictions. So far, the best published results are reported by window-based memory networks.

\vspace{-0.1em}
\tf{bAbI} \cite{weston2016towards} is a collection of artificial datasets, consisting of 20 different reasoning types. It encourages the development of models with the ability to chain reasoning, induction\slash deduction, etc., so that they can answer a question like ``The football is in the \ti{playground}'' after reading a sequence of sentences ``John is in the playground; Bob is in the office; John picked up the football; Bob went to the kitchen.'' Various types of memory networks \cite{sukhbaatar2015end,kumar2016ask} have been shown effective on these tasks, and \newcite{lee2016reasoning} show that vector space models based on extensive problem analysis can obtain near-perfect accuracies on all the categories. Despite these promising results, this dataset is limited to a small vocabulary (only 100--200 words) and simple language variations, so there is still a huge gap from real-world datasets that we need to fill in.

\section{Conclusion}

In this paper, we carefully examined the recent \ti{CNN}\slash \ti{Daily Mail} reading comprehension task. Our systems demonstrated state-of-the-art results, but more importantly, we performed a careful analysis of the dataset by hand.

Overall, we think the \ti{CNN}\slash \ti{Daily Mail} datasets are valuable datasets, which provide a promising avenue for training effective statistical models for reading comprehension tasks. Nevertheless, we argue that: (i)~this dataset is still quite noisy due to its method of data creation and coreference errors; (ii)~current neural networks have almost reached a performance ceiling on this dataset; and (iii)~the required reasoning and inference level of this dataset is still quite simple.

As future work, we need to consider how we can utilize these datasets (and the models trained upon them) to help solve more complex RC reasoning tasks (with less annotated data).

\section*{Acknowledgments}

We thank the anonymous reviewers for their thoughtful feedback. Stanford University gratefully acknowledges the support of the Defense Advanced Research Projects Agency (DARPA) Deep Exploration and Filtering of Text (DEFT) Program under Air Force Research Laboratory (AFRL) contract no. FA8750-13-2-0040. Any opinions, findings, and conclusion or recommendations expressed in this material are those of the authors and do not necessarily reflect the view of the DARPA, AFRL, or the US government.

\bibliography{refs}
\bibliographystyle{acl2016}

\appendix
\section{Samples and Labeled Categories from the \ti{CNN} Dataset}
\label{appendix:samples}
For the analysis in Section \ref{sec:data-analysis}, we uniformly sampled 100 examples from the development set of the \ti{CNN} dataset.  Table \ref{table:index} provides a full index list of our samples and Table \ref{table:categories} presents our labeled categories.

\begin{table}[htp]
    \centering
    \begin{tabular}{p{5cm}|p{8cm}}
    \toprule
    Category & Sample IDs \\
    \midrule
    Exact match (13) & 8, 11, 23, 27, 28, 32, 43, 57, 63, 72, 86, 87, 99 \\
    \midrule
    Sentence-level paraphrasing (41) & 0, 2, 7, 9, 12, 14, 16, 18, 19, 20, 29, 30, 31, 34, 36, 37, 39, 41, 42, 44, 47, 48, 52,
54, 58, 64, 65, 66, 69, 73, 74, 78, 80, 81, 82, 84, 85, 90, 92, 95, 96 \\
    \midrule
    Partial clues (19) & 4, 17, 21, 24, 35, 38, 45, 53, 55, 56, 61, 62, 75, 83, 88, 89, 91, 97, 98 \\
    \midrule
    Multiple sentences (2) & 5, 76 \\
    \midrule
    Coreference errors (8) & 6, 22, 40, 46, 51, 60, 68, 94 \\
    \midrule
    Ambiguous or very hard (17) & 1, 3, 10, 13, 15, 25, 26, 33, 49, 50, 59, 67, 70, 71, 77, 79, 93 \\
    \bottomrule
    \end{tabular}
    \caption{Our labeled categories of the 100 samples.}
    \label{table:categories}
\end{table}

\clearpage
\begin{table*}[!ht]
    \tinyweeny
    \centering
    \scalebox{1.3}{%
    \begin{tabular}{@{}l@{\quad}l|l@{\quad}l@{}}
    \toprule
    ID & Filename & ID & Filename \\
    \midrule
    0 & ddb1e746f88a22fee654ecde8f018e7586595045.question &  1 & 2bef8ec21b10a3294b1496d9a86f29f0592d2300.question \\
2 & 38c702812a874f983e9890c32ba832841a327351.question &  3 & 636857045cf266dd69b67b1e53617bed5253dc33.question \\
4 & 417cbffd5e6275b3c42cb88be222a9f6c7d415f1.question &  5 & ef96409c707a699e4055a1d0684eecdb6e115c16.question \\
6 & b4e157a6a34bf11a03e0b5cd55065c0f39ac8d60.question &  7 & 1d75e7c59978c7c06f3aecaf52bc35b8919eee17.question \\
8 & 223c8e3aeddc3f65fee1964df17bb72f89b723e4.question &  9 & 13d33b8c86375b0f5fdc856116e91a7355c6fc5a.question \\
10 & 378fd418b8ec18dff406be07ec225e6bf53659f5.question &  11 & d8253b7f22662911c19ec4468f81b9db29df1746.question \\
12 & 80529c792d3a368861b404c1ce4d7ad3c12e552a.question &  13 & 728e7b365e941d814676168c78c9c4f38892a550.question \\
14 & 3cf6fb2c0d09927a12add82b4a3f248da740d0de.question &  15 & 04b827f84e60659258e19806afe9f8d10b764db1.question \\
16 & f0abf359d71f7896abd09ff7b3319c70f2ded81e.question &  17 & b6696e0f2166a75fcefbe4f28d0ad06e420eef23.question \\
18 & 881ab3139c34e9d9f29eb11601321a234d096272.question &  19 & 66f5208d62b543ee41accb7a560d63ff40413bac.question \\
20 & f83a70d469fa667f0952959346b496fbf3cdb35c.question &  21 & 1853813a80f83a1661dd3f6695559674c749525e.question \\
22 & 02664d5e3af321afbaf4ee351ba1f24643746451.question &  23 & 20417b5efb836530846ddf677d1bd0bbc831643c.question \\
24 & 42c25a01801228a863c508f9d9e95399ea5f37a4.question &  25 & 70a3ba822770abcaf64dd131c85ec964d172c312.question \\
26 & b6636e525ad58ffdc9a7c18187fb3412660d2cdd.question &  27 & 6147c9f2b3d1cc6fbc57c2137f0356513f49bf46.question \\
28 & 262b855e2f24e1b2e4e0ba01ace81a1f214d729e.question &  29 & d7211f4d21f40461bb59954e53360eeb4bb6c664.question \\
30 & be813e58ae9387a9fdaf771656c8e1122794e515.question &  31 & ad39c5217042f36e4c1458e9397b4a588bbf8cf9.question \\
32 & 9534c3907f1cd917d24a9e4f2afc5b38b82d9fca.question &  33 & 3fbe4bfb721a6e1aa60502089c46240d5c332c05.question \\
34 & 6efa2d6bad587bde65ca22d10eca83cf0176d84f.question &  35 & 436aa25e28d3a026c4fcd658a852b6a24fc6935e.question \\
36 & 0c44d6ef109d33543cfbd26c95c9c3f6fe33a995.question &  37 & 8472b859c5a8d18454644d9acdb5edd1db175eb5.question \\
38 & fb4dd20e0f464423b6407fd0d21cc4384905cf26.question &  39 & a192ddbcecf2b00260ae4c7c3c20df4d5ce47a85.question \\
40 & f7133f844967483519dbf632e2f3fb90c5625a4c.question &  41 & 29b274958eb057e8f1688f02ef8dbc1c6d06c954.question \\
42 & 8ea6ad57c1c5eb1950f50ea47231a5b3f32dd639.question &  43 & 1e43f2349b17dac6d1b3143f8c5556e2257be92c.question \\
44 & 7f11f0b4f6bb9aaa3bdc74bffaed5c869b26be97.question &  45 & 8e6d8d984e51adb5071aad22680419854185eaea.question \\
46 & 57fc2b7ffcfbd1068fbc33b95d5786e2bff24698.question &  47 & 57b773478955811a8077c98840d85af03e1b4f05.question \\
48 & d857700721b5835c3472ba73ef7abfad0c9c499f.question &  49 & f8eedded53c96e0cb98e2e95623714d2737f29da.question \\
50 & 4c488f41622ad48977a60c2283910f15a736417e.question &  51 & 39680fd0bff53f2ca02f632eabbc024d698f979e.question \\
52 & addd9cebe24c96b4a3c8e9a50cd2a57905b6defb.question &  53 & 50317f7a626e23628e4bfd190e987ad5af7d283e.question \\
54 & 3f7ac912a75e4ef7a56987bff37440ffa14770c6.question &  55 & 610012ef561027623f4b4e3b8310c1c41dc819cc.question \\
56 & d9c2e9bfc71045be2ecd959676016599e4637ed1.question &  57 & 848c068db210e0b255f83c4f8b01d2d421fb9c94.question \\
58 & f5c2753703b66d26f43bafe7f157803dc96eedbc.question &  59 & 4f76379f1c7b1d4acc5a4c82ced64af6313698dd.question \\
60 & e5bb1c27d07f1591929bf0283075ad1bc1fc0b50.question &  61 & 33b911f9074c80eb18a57f657ad01393582059be.question \\
62 & 58c4c046654af52a3cb8f6890411a41c0dd0063b.question &  63 & 7b03f730fda1b247e9f124b692e3298859785ef3.question \\
64 & ece6f4e047856d5a84811a67ac9780d48044e69a.question &  65 & 35565dc6aecc0f1203842ef13aede0a14a8cf075.question \\
66 & ddf3f2b06353fe8a9b50043f926eb3ab318e91b2.question &  67 & e248e59739c9c013a2b1b7385d881e0f879b341d.question \\
68 & e86d3fa2a74625620bcae0003dfbe13416ee29cf.question &  69 & 176bf03c9c19951a8ae5197505a568454a6d4526.question \\
70 & ee694cb968ae99aea36f910355bf73da417274c0.question &  71 & 7a666f78590edbaf7c4d73c4ea641c545295a513.question \\
72 & 91e3cdd46a70d6dfbe917c6241eab907da4b1562.question &  73 & e54d9bdcb478ecc490608459d3405571979ef3f2.question \\
74 & f3737e4de9864f083d6697293be650e02505768c.question &  75 & 1fc7488755d24696a4ed1aabc0a21b8b9755d8c6.question \\
76 & fb3eadd07b9f1df1f8a7a6b136ad6d06f4981442.question &  77 & 1406bdad74b3f932342718d5d5d0946a906d73e2.question \\
78 & 54b6396669bdb2e30715085745d4f98d058269ef.question &  79 & 0a53102673f2bebc36ce74bf71db1b42a0187052.question \\
80 & d5eb4f98551d23810bfeb0e5b8a94037bcf58b0d.question &  81 & 370de4ffe0f2f9691e4bd456ff344a6a337e0edf.question \\
82 & 12f32c770c86083ff21b25de7626505c06440018.question &  83 & 9f6b5cff3ce146e21e323a1462c3eff8fca3d4a0.question \\
84 & 1c2a14f525fa3802b8da52aebaa9abd2091f9215.question &  85 & f2416e14d89d40562284ba2d15f7d5cc59c7e602.question \\
86 & adcf5881856bcbaf1ad93d06a3c5431f6a0319ba.question &  87 & 097d34b804c4c052591984d51444c4a97a3c41ac.question \\
88 & 773066c39bb3b593f676caf03f7e7370a8cd2a43.question &  89 & 598cf5ff08ea75dcedda31ac1300e49cdf90893a.question \\
90 & b66ebaaefb844f1216fd3d28eb160b08f42cde62.question &  91 & 535a44842decdc23c11bae50d9393b923897187e.question \\
92 & e27ca3104a596171940db8501c4868ed2fbc8cea.question &  93 & bb07799b4193cffa90792f92a8c14d591754a7f3.question \\
94 & 83ff109c6ccd512abdf317220337b98ef551d94a.question &  95 & 5ede07a1e4ac56a0155d852df0f5bb6bde3cb507.question \\
96 & 7a2a9a7fbb44b0e51512c61502ce2292170400c1.question &  97 & 9dcdc052682b041cdbf2fadc8e55f1bafc88fe61.question \\
98 & 0c2e28b7f373f29f3796d29047556766cc1dd709.question &  99 & 2bdf1696bfd2579bb719402e9a6fa99cb8dbf587.question \\
	\bottomrule
	\end{tabular}
	} 
	\caption{A full index list of our samples.}
	\label{table:index}
\end{table*}

\end{document}